\documentclass[runningheads]{llncs}
\usepackage{graphicx}
\usepackage[utf8]{inputenc}
\usepackage{algorithmic}
\usepackage{textcomp}
\usepackage{xcolor}
\usepackage{booktabs}
\usepackage{subcaption}
\usepackage{flushend}
\usepackage{linguex}
\usepackage{tcolorbox}
\definecolor{fhgreen}{RGB}{0,147,116}
\newtcolorbox{mybox}{fontupper=\footnotesize}
\usepackage{amsmath}
\usepackage{comment}
\renewcommand\thefootnote{*}
\newenvironment{hide}{}{}
\excludecomment{hide}
\begin{document}
\title{Towards Linguistically Informed Multi-Objective Transformer Pre-Training for Natural Language Inference$^*$}

%
%
\author{Maren Pielka\inst{1} \and
Svetlana Schmidt\inst{1,2} \and
Lisa Pucknat\inst{1,3} \and
Rafet Sifa\inst{1}}
\authorrunning{M. Pielka et al.}
%
\institute{Fraunhofer IAIS, Schloss Birlinghoven, 53757 Sankt Augustin, Germany \and
Ruhr-Universität Bochum, Universitätsstraße 150, 44801 Bochum, Germany \and
Rheinische Friedrich-Wilhelms Universität Bonn, Regina-Pacis-Weg 3, 53113 Bonn, Germany
}
\maketitle\footnotetext[1]{This work was accepted in the proceedings of ECIR 2023. The copyright lies with ACM.}   
\addtocounter{footnote}{-1}
\renewcommand\thefootnote{\arabic{footnote}}

\begin{abstract}
We introduce a linguistically enhanced combination of pre-training methods for transformers. The pre-training objectives include POS-tagging, synset prediction based on semantic knowledge graphs, and parent prediction based on dependency parse trees. Our approach achieves competitive results on the Natural Language Inference task, compared to the state of the art. Specifically for smaller models, the method results in a significant performance boost, emphasizing the fact that intelligent pre-training can make up for fewer parameters and help building more efficient models. Combining POS-tagging and synset prediction yields the overall best results.
\end{abstract}

\section{Introduction}
Understanding entailment and contradictions is a particularly hard task for any machine learning (ML) model. The system has to deeply comprehend the semantics of natural language, and have access to some amount of background knowledge that is often helpful in understanding many real-world statements. Current solutions still show deficits with respect to both criteria. At the same time, state-of-the-art language models such as GPT-3 and XLM-RoBERTa rely heavily on massive amounts of data for pre-training, and are quite resource-extensive due to their large number of parameters. 
        
To address those shortcomings, we present a linguistically enhanced approach for multi-objective pre-training of transformer models. We inject extra knowledge into the model by pre-training for three additional language modelling tasks, one of which is a novel approach. Specifically, we utilize external information about part of speech tags, syntactic parsing, and semantic relations between words. Our main contribution can be summarized as follows:

\vspace{2mm}
\begingroup
\leftskip4em
\rightskip\leftskip
We aim to become independent of huge data resources for pre-training, and having to train models with a large number of parameters, by injecting as much external knowledge to the model as possible. This goal is being quantified by evaluating our model on the Stanford Natural Language Inference (SNLI) \cite{Bowman2015-EMNLP} data set. We compare different implementations of the transformer architecture (BERT \cite{devlin2018bert} and XLM-RoBERTa \cite{liu_roberta,conneau2019unsupervised}), with the aim to show that the smaller model, BERT, is able to perform competitively when being enhanced with additional knowledge during pre-training.
\par
\endgroup
\vspace{2mm}
Our approach does not require any additional data for pre-training, it is pre-trained for the additional tasks on the same data set that it is later being fine-tuned on.

\section{Related Work}
In Natural Language Inference (NLI), first introduced by \cite{dagan2005pascal}, one has to determine whether a given \textit{hypothesis} can be inferred from a given \textit{premise}, or whether it contradicts the premise. 
Further, a new research field emerged from the NLI task named Contradiction Detection (CD). Multiple languages, besides the commonly used English language, such as Persian \cite{rahimi2021contradiction}, Spanish \cite{sepulveda2021here}, and German \cite{sifa2019towards,cidm2019,icpr2021,pucknat2021detecting} were studied. We follow up on the latter research, in which a portion of the SNLI dataset was machine-translated into German. They found that RNNs handle machine-translated data quite well, with difficulties in complicated sentence structures, translation artifacts, and exploitation of world knowledge. A fine-tuned XLM-RoBERTa model seemed to be most promising with regard to the difficulties mentioned above. Still, qualitative exploration has shown that among other things, the model struggles with prepositional references, incomplete sentences as well as antonyms and homonyms, which gave rise to enhance the model with lower level linguistic tasks.

BERT \cite{devlin2018bert} and XLM-RoBERTa \cite{liu_roberta,conneau2019unsupervised} are among the state of the art transformer-based encoder models for text classification tasks. They use the pre-training objectives of Masked Language Modeling and Next Sentence Prediction (only BERT) to obtain a large amount of language understanding in an unsupervised way.
Dependency Injected Bidirectional Encoder Representations from Transformers (DIBERT) \cite{wahabcidm2021} utilizes a third pre-training objective called parent prediction injecting syntactic structures of dependency trees. 

The approach of integrating the external semantic knowledge into a transformer model was presented by \cite{barbouch2021wn}. In their work WordNet embeddings are combined with the BERT architecture in two ways: during \textit{external combination} the outputs of WordNet and BERT are combined for the additional classification and in \textit{internal inclusion} the WordNet representations are integrated into the internal BERT architecture. The resulting models were evaluated on four GLUE \cite{wang2018glue} datasets for Sentiment Analysis, Linguistic Acceptability, Sentence Similarity and Natural Language Inference tasks \cite{barbouch2021wn}. 

A similar approach has been introduced by \cite{limit_bert2019}. They pre-train a BERT model on five different, linguistics aware tasks such as POS-tagging, semantic role labeling and syntactic parsing, achieving competitive results on GLUE benchmark tasks. The main difference between this work and ours is that we focus on minimizing the amount of pre-training data and model parameters, utilizing the same data sets for both custom pre-training and fine-tuning. In addition, we introduce the novel synset\footnote{https://wordnet.princeton.edu/} prediction objective. Unlike the approach of \cite{barbouch2021wn}, we utilize only one synset extracted for each word in the data.


\section{Data}
The Stanford Natural Language Inference (SNLI) data set was introduced by Bowman et al. \cite{Bowman2015-EMNLP}. It is the largest collection of human-generated premise and hypothesis pairs for the NLI task with over 570,000 examples. The data was collected in a crowd-source campaign. Workers were instructed to devise hypotheses inspired by premise image captions. These sentences should entail, contradict or not relate to the original caption. In a final effort, sentence pairs were labeled by different annotators with one of three labels - \textit{entailment}, \textit{contradiction} or \textit{neutral} (if hypothesis do not relate to premise). The gold label for each pair was chosen based on a majority vote of annotators.

\begin{hide}
\begin{table*}[h!]
\small
    \centering
    \resizebox{\textwidth}{!}{
    \begin{tabular}{cccc} \toprule
        & \textbf{Premise} & \textbf{Hypothesis} & \textbf{Label} \\ \midrule
        \textbf{1} & \begin{minipage}{5cm} \vspace{1mm} "Die Qualität der Kette ist sehr gut. Die Kette sieht hochwertig aus und die Lieferung war wirklich schnell :)" - \textit{``The quality of the necklace is very good. The necklace looks high-quality and delivery was really fast :)"} \vspace{1mm} \end{minipage} & \begin{minipage}{5cm} \vspace{1mm} "Keine Benachrichtigung über Sendung und keine sendungsverfolgung möglich.  Zu lange Lieferzeiten."- \textit{``No notification on the shipment and shipment tracking not possible. Delivery times too long."} \vspace{1mm} \end{minipage} & "contradiction" \\ \midrule
         \textbf{2} & \begin{minipage}{5cm} \vspace{1mm} "Das Spiel bietet mit seinen vier Klassen jede Menge Spielspaß, ob allein, oder im Koop." - \textit{``The game offers a lot of gaming fun with its four classes, whether you're alone or in coop mode."} \vspace{1mm} \end{minipage} & \begin{minipage}{5cm} \vspace{1mm} "Ausgeklügeltes Spielsystem sorgt für Spannung." - \textit{``Sophisticated gaming system ensures suspense."} \vspace{1mm} \end{minipage} & "no contradiction" \\ 
         
          \bottomrule
    \end{tabular} }
    \caption{Example pairs with labels from the internet data set, original German version and English translation (in italic)}
    \label{tab:internet_examples}
\end{table*}
\end{hide}

\section{Methodology: Pretraining methods}
Injecting syntactic and semantic information into the architecture is achieved by training with different pre-training objectives. 
All of our pre-training objectives are word-based, meaning that we utilize the output vector mapping to the corresponding input-token for these tasks instead of the special [CLS] token, which is commonly used for classification tasks. 
All of our labels are generated in a semi-supervised manner. We take advantage of already present and well working architectures to predict labels for POS-tagging and dependency parsing and create labels for different synsets with nltk wordnet\footnote{https://www.nltk.org/howto/wordnet.html} interface supporting WordNet\footnote{https://wordnet.princeton.edu/} lexical database.


\subsection{POS-tagging}
The main objective of part-of-speech (POS)-tagging is to predict the syntactic function of a word in a sentence. Words can have different meanings in different contexts. Therefore, POS tagging is used, among other things, to identify the context in which a word occurs. The used tagset includes common parts of speech such as adjective, noun and verb, but also finer graduations such as numerical and symbol words and grammatical tags such as adverb and pronoun. The following example shows semi-supervised generated tags for a tokenized sentence from the SNLI dataset. An underscore corresponds to the beginning of a word. Each input-token is assigned the POS-tag for the complete word. Tokens for words that are split up by the tokenizer map to the same POS-tag.

\scriptsize
\newcommand{\undset}[2]{\underset{\scriptscriptstyle #1}{#2\strut}}
\begin{tcolorbox}[notitle,boxrule=0pt,boxsep=0pt,colback=gray!10,colframe=gray!10]
$\undset{\text{{\color{fhgreen}{DET}}}}{\text{{\_A }}}
\undset{\text{{\color{fhgreen}{NOUN}}}}{\text{{\_person }}}
\undset{\text{{\color{fhgreen}{ADP}}}}{\text{{\_on\quad}}}
\undset{\text{{\color{fhgreen}{DET}}}}{\text{{\_a }}}
\undset{\text{{\color{fhgreen}{NOUN}}}}{\text{{\_horse }}}
\undset{\text{{\color{fhgreen}{VERB}}}}{\text{{\_jump }}}
\undset{\text{{\color{fhgreen}{VERB}}}}{\text{{s }}}
\undset{\text{{\color{fhgreen}{ADP}}}}{\text{{\_over }}}
\undset{\text{{\color{fhgreen}{DET}}}}{\text{{\_a }}}
\undset{\text{{\color{fhgreen}{VERB}}}}{\text{{\_broken }}}
\undset{\text{{\color{fhgreen}{ADP}}}}{\text{{\_down }}}
\undset{\text{{\color{fhgreen}{NOUN}}}}{\text{{\_air\quad}}}
\undset{\text{{\color{fhgreen}{NOUN}}}}{\text{{plan\quad}}}
\undset{\text{{\color{fhgreen}{NOUN}}}}{\text{{e  }}}
\undset{\text{{\color{fhgreen}{ PUNCT}}}}{\text{{  .}}}$
\label{pos_example}
\end{tcolorbox}

\normalsize
We extract labels from spaCys implementation for POS-tagging \cite{honnibal2020spacy}. 

\subsection{Parent prediction}
For parent prediction (PP) \cite{wahabcidm2021} the parent of each word is predicted. The parent is deduced from a corresponding dependency tree of the sentence, which was created using the NLP library Stanza \cite{qi2020stanza}. The dependency tree provides information about the syntactic dependency relation between words. Each word is assigned to exactly one other word, so each word has precisely one parent. The central clause, i.e. the root clause without parent, is a (finite) verb. 

\subsection{Synset prediction}

In order to enhance the model with semantic knowledge, we take advantage of the WordNet \cite{miller1995wordnet,wordnet} knowledge graph. WordNet is the lexical database for the English language. The nouns, verbs, adjectives and adverbs in WordNet are organized in groups, based on their semantic similarity. Those groups are called synsets (synonyms sets). 

\ex. \label{first_syn} 
The lady is weeding her garden.

One synset represents one distinct concept, thus one synset can contain several lexical units, where each of the lexical units represents one meaning of a word. Since words have several meanings, they can be associated with several synsets. For instance, the synset for the word \textit{lady} in \ref{first_syn} contains three possible meanings, as it can be seen below.

\begin{tcolorbox}[notitle,boxrule=0pt,boxsep=0pt,colback=gray!10,colframe=gray!10]\label{sec_syn}
\color{fhgreen}{Synset('lady.n.01'), Synset('dame.n.02'), Synset('lady.n.03')}
\end{tcolorbox}

The main objective of this pre-training task is the prediction of labels representing semantic knowledge. We extract the synsets from WordNet for nouns, verbs and adjectives. The WordNet\footnote{https://www.nltk.org/\_modules/nltk/corpus/reader/wordnet.html} nltk corpus reader is used for the extraction of synsets. The first synset in a set of synsets represents the most common meaning of a word. Thus, we utilize the first synset for semantic representation of a word. For example, for the word \textit{lady} the synset \textit{Synset('lady.n.01')} is chosen. We argue that since most words have a unique meaning, this approach is a reasonable heuristic, even though it will introduce a small amount of noise by assigning the wrong synset to uncommon words.

The following example \ref{ex_syn} shows the tokenized sentence \textit{"The lady is weeding her garden."} and the corresponding labels. Similar to the example in \ref{pos_example} the label for a complete token is assigned to each of the subtokens, just as in case with \textit{\_we ed ing}.

\begin{tcolorbox}[notitle,boxrule=0pt,boxsep=0pt,colback=gray!10,colframe=gray!10]
$\undset{\text{{\color{fhgreen}{no\_syn}}}}{\text{{\_The }}}
\undset{\text{{\color{fhgreen}{lady\_n\_01}}}}{\text{{\_lady }}}
\undset{\text{{\color{fhgreen}{be\_v\_01}}}}{\text{{\_is}}}
\undset{\text{{\color{fhgreen}{weed\_v\_01 }}}}{\text{{\_we }}}
\undset{\text{{\color{fhgreen}{weed\_v\_01 }}}}{\text{{ed }}}
\undset{\text{{\color{fhgreen}{weed\_v\_01 }}}}{\text{{ing  }}}
\undset{\text{{\color{fhgreen}{no\_syn}}}}{\text{{\_her }}}
\undset{\text{{\color{fhgreen}{garden\_n\_01}}}}{\text{{\_garden }}}$
\label{ex_syn}
\end{tcolorbox}

To our best knowledge, it is the first attempt to utilise the synsets for pre-training the model with semantic knowledge.

\section{Experiments and Results}
In the next section, we describe the experimental setup and further evaluate our proposed pre-training objectives quantitatively and qualitatively. 
We do not use any additional data, other than the SNLI training set, and prolong the overall training only by a few epochs, since the same data is used with a different objective. 

The main model\footnote{\url{https://huggingface.co/bert-base-cased}} is based on a BERT architecture with approximately 110M parameters with 12-layers, 12 attention heads and a hidden state of size 768. A simple feed-forward layer is used for classification and shared across each output vector or, in case of finetuning, for the special [CLS] token. The BERT model has been pre-trained for the Masked Language Modeling and Next Sentence Prediction tasks on a large corpus of English data from books \cite{zhuBooks} and Wikipedia. Further, we compare to a large XLM-RoBERTa\footnote{\url{https://huggingface.co/xlm-roberta-large}} with 355M parameters. Binary Cross Entropy Loss in combination with AdamW optimizer \cite{loshchilov2017decoupled} is used for all experiments. For pre-training a learning rate of 6e-5 is used. For fine-tuning we utilized a learning rate of 5e-6. 

Evaluating our main model, the overall best results are achieved when we pre-train for POS-tagging and synset prediction, yielding a significant performance boost over the baseline model (see table \ref{tab:comparison_three_way}). This proofs that linguistically informed pre-training does in fact help the model to capture additional knowledge that is helpful for the classification task. Apparently, not all combinations of pre-training methods work equally well. For example, combining all three approaches yields slightly worse results than combining only POS-tagging and parent prediction, or POS-tagging and synset prediction. 

Comparing the different model architectures (table \ref{tab:comparison_model_sizes}), it is apparent that adding further pre-training tasks helps the smaller models achieve competitive results compared to xlm-roberta-large, while it does not yield a huge performance boost for the large model itself. This suggests that enhancing the smaller models with additional knowledge could make them competitive, and thereby not having to rely on extensive computational resources. At the same time, both models achieve results that are comparable to the current state of the art \cite{wang2021entailment,sun2020self-explaining}.

\begin{hide}
\subsection{Qualitative Evaluation}
For chosen pre-trained models, we select example sentences to evaluate the predicted labels on. They are depicted in Table \ref{tab:qualitative_exploration}. To begin with, it should be pointed out that all three models considered achieve very good accuracy results overall. Still, there are examples that might indicate weaknesses in the models abilities. Considering the first pair of examples in table \ref{tab:qualitative_exploration}, it is possible that the standard model without additional training draws incorrect conclusions about the affiliation of the color. Lacking affiliation might also arise in example four. Next, it is interesting to notice that sentence pair five is wrongly classified by all models, which hints to the fact that similar sentences could lead to similar sentence representations and are therefor harder to classify correctly.

{
\begin{table*}[h]
    \centering
     \footnotesize
   \begin{subtable}[h]{0.45\textwidth}
   \centering
   \begin{tabular}{llc}
    \toprule 
    \textbf{Pretraining Conf.}& \textbf{Stacking Approach} & \textbf{Acc.}  \\ \midrule
         No add. pretraining & No stacking & 96.6 \\
         POS & No stacking & 96.2  \\
         POS & POS-0 & 96.6 \\
         PP & No stacking & 96.4 \\
         PP & PP-0 & 96.6 \\
         POS+PP & No stacking & 95.4 \\
         Hypernyms & No stacking & 96.4 \\
         POS+Hypernyms & No stacking &  \\
         POS+PP+Hypernyms & No stacking&  \\
         PP+Hypernyms & No stacking &  \\
         POS+PP+Hypernyms & No stacking &  \\
    \bottomrule
    \end{tabular} 
    \caption{Results on the original SNLI test set}
   \label{tab:comparison_snli_orig}
    \vspace{2mm}
   \end{subtable}
   \hfill
    \begin{subtable}[h]{0.45\textwidth}
    \centering
   \begin{tabular}{llc}
    \toprule 
    \textbf{Pretraining Conf.}& \textbf{Stacking Approach} & \textbf{Acc.} \\ \midrule
         No add. pretraining & No stacking & 92.4 \\
         POS & No stacking & 93.3 \\
         POS & POS-0 & 93.4 \\
         PP & No stacking & 92.6 \\
         PP & PP-0 & 93.2\\
         POS+PP & No stacking & 92.3 \\
         POS+PP & POS-0, PP-1 \\
         Syn & No stacking & 93.0 \\
         Syn & Syn-0 & 93.0\\
         POS+Syn & POS-0, Syn-1 & 93.2\\
         POS+PP+Syn & POS-0, PP-1, Syn-2 \\
    \bottomrule
    \end{tabular} 
    \caption{Results on the translated SNLI test set}
    \label{tab:comparison_snli_trans}
    \vspace{2mm}
     \end{subtable}
    
    \begin{subtable}[h]{\textwidth}
    \centering
   \begin{tabular}{llc}
    \toprule 
    \textbf{Pretraining Conf.}& \textbf{Stacking Approach} & \textbf{Acc.} \\ \midrule
         No add. pretraining & No stacking & 89.4\\
         POS & No stacking & 86.7 \\
         POS & POS-0 & 88.7 \\
         PP & No stacking & 89.2  \\
         PP & PP-0 & \textbf{89.5} \\
         POS+PP & No stacking & 87.2 \\
         POS+PP & POS-0, PP-1 & \\
         Syn & No stacking \\
         Syn & Syn-0 & 89.4 \\
         POS+Syn & POS-0, Syn-1 & 88.6 \\
         POS+PP+Syn & POS-0, PP-1, Syn-2 \\
    \bottomrule
    \end{tabular} 
    \caption{Results on the German internet data set}
    \label{tab:comparison_snli_trans}
    \vspace{2mm}
     \end{subtable}
    \caption{Performance comparison for different pre-training and stacking configurations on the two-way task (contradiction vs. no contradiction), in percent. The abbreviations stand for: POS=POS-tagging, PP=Parent Prediction, Syn=Synset prediction. The numbers in the stacking approaches encode the transformer stacks. So "POS-0, PP-1, Syn-2" means "Training stack 0 for POS-tagging, stack 1 for Parent Prediction, and stack 2 for Synset Prediction.}
\label{tab:comparison_two_way}
\end{table*}{}}
\end{hide}

{
\begin{table*}[h]
    \centering
    \resizebox{\textwidth}{!}{
     \footnotesize
   \begin{tabular}{lcccc}
    \toprule 
    \textbf{Pretraining Configuration} & \textbf{Acc.} & \textbf{F1-Score (Cont.)}& \textbf{F1-Score (Ent.)} & \textbf{F1-Score (Neut.)}  \\ \midrule
         No additional pretraining & 88.6 & 91.6 & 89.7 & 84.5 \\
         POS & 90.0 & 92.4 & 90.9 & 86.7 \\
         PP & 89.5 & 92.1 & 90.4 & 85.9 \\
         POS+PP & 90.2 & 92.8 & 91.1 & 86.5 \\
         Syn & 89.9 & 92.3 & 90.8 & 86.6 \\
         POS+Syn & \textbf{90.4} & \textbf{93.2} & \textbf{91.7} & \textbf{86.7} \\
         PP+Syn & 89.9 & 92.6 & 90.6 & 86.3 \\
         POS+PP+Syn & 89.9 & 92.5 & 90.7 & 86.4 \\
    \bottomrule
    \end{tabular} }
   \label{tab:comparison_snli_orig}
    \vspace{3mm}

    \caption{Performance comparison for different pre-training configurations on the SNLI test set, in percent. The abbreviations stand for: POS=POS-tagging, PP=Parent Prediction, Syn=Synset prediction.}
\label{tab:comparison_three_way}
\end{table*}{}}

{
\begin{table*}[h]
    \centering
   \resizebox{\textwidth}{!}{
   \begin{tabular}{lllcccc}
    \toprule 
    \textbf{Configuration}& \textbf{Base model} & \textbf{Num. param.}& \textbf{Acc.} & \textbf{F1 (Cont.)}& \textbf{F1 (Ent.)} & \textbf{F1 (Neut.)}  \\ \midrule
        Current SOTA (EFL) & roberta-large & 355 M & 93.1 & n.a. & n.a. & n.a \\
         No add. pretraining & xlm-roberta-large & 345 M & 91.5 & 94.5 & 92.1 & 87.7 \\
         POS+Syn & xlm-roberta-large & 345 M &  91.5 & 94.5 & 92.0 & 88.1 \\
         No add. pretraining & bert-base-cased & 110 M & 88.6 & 91.6 & 89.7 & 84.5 \\
         POS+Syn & bert-base-cased & 110 M & 90.4 & 93.2 & 91.1 & 86.7 \\
    \bottomrule
    \end{tabular} }
    \vspace{3mm}

    \caption{Performance comparison for different model architectures and sizes on the SNLI test set, in percent. We compare our approaches with (POS+Syn) and without pre-training to the current best result on the data set by \cite{wang2021entailment}.}
\label{tab:comparison_model_sizes}
\end{table*}{}}

\begin{hide}

{
\begin{table*}
\centering
\footnotesize
\resizebox{\textwidth}{!}{
\begin{tabular}{p{0.3cm}p{5cm}p{3.5cm}cccc}
\toprule
& & & \multicolumn{4}{c}{\textbf{Label}} \\  \cmidrule{4-7}
& \textbf{Premise} &
  \textbf{Hypothesis} &
  \textbf{Gold} &
  \textbf{No Pre} &
  \textbf{POS-0} &
  \textbf{PP-0} \\ \midrule
  \textbf{1} &
Two boys, one wearing green, the other in yellow, are playing with sidewalk chalk, drawing pictures on the ground. &
  Two boys are drawing pictures with pink chalk. &
  \omark &
  \xmark &
  \omark &
  \omark \\ \midrule
 \textbf{2} &
A man in swimming trunks plays in the ocean's waves. &
  A man is surfing out on the ocean. &
  \omark &
  \xmark &
  \xmark &
  \omark \\\midrule
  \textbf{3} &
A man walks down the alley talking on his cellphone. &
  A man is walking down a wooded path and talking on the phone. &
  \xmark &
  \omark &
  \xmark &
  \omark \\\midrule
  \textbf{4} &
A soccer game in a large area with 8 yellow players and 4 black players. &
  There is a soccer game with 12 players. &
  \cmark &
  \xmark &
  \xmark &
  \cmark \\\midrule
  \textbf{5} &
Men and women in swimsuits hangout on rocks above water. &
  The rocks hangout above the water. &
  \xmark &
  \cmark &
  \cmark &
  \cmark \\
  \bottomrule
\end{tabular}}
\caption{Legend: \xmark = Contradiction, \cmark = Entailment, \omark = Neutral \\ 
Exemplary sentences from the SNLI dataset. Gold label refers to the actual label assigned by annotators, and No Pre, POS-0, and PP-0 correspond to the predicted label given by the models without further pre-training, pre-training of the first encoder stack with POS tagging, and pre-training of the first encoder stack with parent prediction.}
\label{tab:qualitative_exploration}

\end{table*}{}}

\end{hide}

\section{Conclusion and Outlook}
We presented a combination of linguistically enhanced  pre-training methods for transformers. The experimental results illustrate that the performance of the transformer models on the NLI task can be improved due to enhancing the models with syntactic and semantic knowledge. The novel method of synset prediction shows that enriching transformer models with semantic knowledge positively affects the ability of the models to learn semantic correlations in data. Moreover, it is not required to utilize a large transformer model for handling the task of detecting contradictions, entailments or neutral expressions. Another important withdrawal of our approach is that the improvement can be achieved with no additional training data.

Future work includes extending the approach to other languages such as German, Italian or Arabic. We also aim to investigate, how we can incorporate even more external knowledge and thereby reduce the model size further. One idea would be data augmentation methods with the goal to align the languages in feature space. Another direction of research is training on prototypical examples, as suggested by \cite{von2022informed}. Those could be created using linguistic rules, thus reducing the amount of hand-annotated training data and teaching the model the essential rules of contradiction and entailment. Finally, we also plan to apply the pre-training approach to other tasks, such as toxicity detection or relation extraction from financial documents.

\section{Acknowledgments}
This research has been funded by the Federal Ministry of Education and Research of Germany and the state of North-Rhine Westphalia as part of the Lamarr-Institute for Machine Learning and Artificial Intelligence, LAMARR22B.
\bibliographystyle{splncs04}
\bibliography{mybibliography}

\end{document}